\documentclass[preprint,12pt]{elsarticle}%

\usepackage{amssymb}

\newcommand{\Levothyrox}{Levothyrox\textsuperscript{\textregistered}~}
\newcommand{\Doctissimo}{Doctissimo\textsuperscript{\textregistered}~}
\usepackage[utf8]{inputenc}
\usepackage{natbib}
\usepackage{graphicx}
\usepackage[table,xcdraw]{xcolor}
\usepackage{booktabs}
\usepackage{svg}

\journal{Artificial Intelligence in Medicine}

\begin{document}

\begin{frontmatter}



\title{AI-based Approach for Safety Signals Detection from Social Networks: Application to the \Levothyrox Scandal in 2017 on \Doctissimo Forum}

\author[inst1]{Valentin Roche}
\affiliation[inst1]{organization={Universit\'e Claude Bernard - Lyon 1, Facult\'e de Pharmacie, Institut des Sciences Pharmaceutiques et Biologiques},
            addressline={8 Avenue Rockefeller},
            postcode={69008},
            city={Lyon},
            country={France}}

\author[inst1]{Jean-Philippe Robert}

\author[label1]{Hanan Salam}
\affiliation[label1]{organization={SMART Lab, New York University Abu Dhabi},
            addressline={Saadiyat Island},
            city={ Abu Dhabi},
            postcode={PO Box 129188},
            state={},
            country={United Arab Emirates}}

\begin{abstract}
    Social media can be an important source of information facilitating the detection of new safety signals in pharmacovigilance. Various approaches have investigated the analysis of social media data using Artificial Intelligence such as Natural Language Processing (NLP) techniques for detecting adverse drug events. Existing approaches have focused on the extraction and identification of Adverse Drug Reactions (ADR), Drug-Drug Interactions (DDI) and drug misuse. However, non of the works tackled the detection of potential safety signals by taking into account the evolution in time of relevant indicators. 
    Moreover, despite the success of deep learning in various healthcare applications, it was not explored for this task. 
    We propose an Artificial Intelligence (AI) based approach for the detection of potential pharmaceutical safety signals from patients' reviews that can be used as part of the pharmacovigilance surveillance process to flag the necessity of immediate attention and encourage the conduct of an in-depth pharmacovigilance investigation.
    We focus on the \Levothyrox case in France which triggered a huge attention from the media following the change of the medication formula, leading to an increase in the frequency of adverse drug reactions normally reported by patients. 
   Our approach is two-fold. (1) We investigate various NLP-based indicators extracted from patients' reviews including words and n-grams frequency, semantic similarity, Adverse Drug Reactions mentions, and sentiment analysis. 
    (2) We propose a deep learning  architecture, named Word Cloud Convolutional Neural Network (WC-CNN) which trains a CNN on word clouds extracted from the patients comments. We study the effect of different time resolutions (day, week, month, combined) as well as different NLP pre-processing techniques on the performance of the model. 
    Our results  show that the proposed indicators could be used in the future to effectively detect new safety signals. Moreover, the deep learning model trained on word clouds extracted at monthly resolution outperform models trained on data extracted at daily, weekly, and combined time resolutions with an accuracy of 75 \%. 
    
\end{abstract}

\begin{graphicalabstract}
\includegraphics{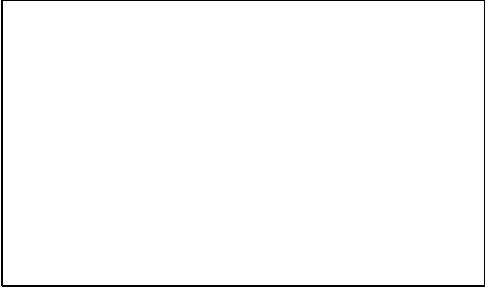}
\end{graphicalabstract}

\begin{highlights}
\item A new Deep Learning approach for the early detection of pharmaceutical safety signals.
\item  Investigate word clouds extracted at different resolutions for potential pharmaceutical safety signals detection. 
\item First approach to use word clouds as input of CNN for classification. 
\item Analysis of patients reviews in medical forums for the early detection of pharmaceutical safety signals.
\item First analysis of the \Levothyrox scandal in 2017 in France, using natural language processing and deep learning tools. 
\item Patients comments in French on \Levothyrox Adverse Drug Reactions following the change of formula were analyzed. 
  \item Application of the proposed approach on a real world use case. 

\end{highlights}


\end{frontmatter}



\section{Introduction}

Pharmacovigilance science refers to the collection, detection, assessment, monitoring, and prevention of Adverse Events (AEs) with pharmaceutical products \cite{world2002importance}. 
A safety signal in Pharmacovigilance is defined to be an information of increasing or decreasing frequency regarding a new or known AE that might be caused by a medicine and requires attention and further investigation  \cite{leadverse} \cite{hauben2009defining}. Safety signals can be detected via a wide range of sources. 
Normally, the signals are detected and sometimes reported by primary care teams (main entry point into the pharmacovigilance system). However, it is common for  patients not to express their feelings to the care teams (white coat effect \cite{pickering2002white}). This phenomenon diminishes once they regain their comfort zone. 

Real-life data, as opposed to data generated from classic experimental settings (randomized controlled trials, double-blind versus reference or placebo), are generated by the daily environment of the patient. Such data is much more frequently and spontaneously exchanged with family members, friends and communities of interest through different channels (orally, by messages, or via social networks). 
The analysis of real-life data has become essential and represents a major challenge for improving the quality of healthcare as well as the regulation of  health systems \cite{murdoch2013inevitable}. 
Such analysis has multiple interests \cite{bate2018hope} and occurs at several stages of the life of a drug; at the level of: (1) medico-economic evaluations and (2) ``Post-Marketing Surveillance'' (phase IV) of pharmacovigilance. In particular, data posted on social media platforms can be an important source for new signals identification allowing patients and healthcare professionals to communicate AEs and Adverse Drug Reactions (ADRs) in a spontaneous and honest manner together while promoting patients exchanges and feedback \cite{european2017guideline}.
For instance specialized social networks such as Doctissimo\textsuperscript{\textregistered}~collect today significant amounts of information, the study of which has become crucial for biomedical research and health systems.

The rise of Artificial Intelligence (AI) and the advances in big data have resulted in the development of new tools to collect, analyze and cross-reference large amounts of data. Consequently, this allowed the exploitation of patients' social media data. Specifically, Natural Language Processing (NLP) and Machine Learning (ML) techniques have been successfully exploited in the literature for the analysis of medical forums for various applications in the medical field such as Adverse Drug Reaction (ADR) identification and extraction  \cite{lardon2015adverse,el2019adverse,arnoux2019adverse}, targeted medical information retrieval \cite{bekhuis2011using}, drug non-compliance detection \cite{bigeard2019detection}, and patients opinion mining and sentiment analysis \cite{jimenez2019we}. 
Furthermore, the rise of deep learning techniques had led researchers to explore different deep architectures and methods for mining patients' reviews \cite{rivas2018automatic}, and detecting ADRs \cite{cocos2017deep,lee2019machine,fan2020adverse}. Compared to traditional Machine Learning approaches, deep neural networks have shown significant improvement in the performance of such tasks due to their capacity to automatically learn high level abstractions from the patient's reviews \cite{lee2021prediction}. Despite the use of deep learning methods for ADE detection and extraction, there has been just a few works that explored deep learning for pharmacovigilance safety signals detection. Most of the existing approaches concentrate on the identification and extraction of ADRs. This work differs from the others by the fact that it concentrates on identifying the occurrence of a possible safety signal that requires immediate attention and further investigation.

In this paper, we propose  a method for the early detection of safety signals detection based on the frequency of adverse events reported by patients on social networks. We focus on the \Levothyrox~case \cite{casassus2018risks,concordet2020were,nicolas2020comment}.
In France, this case received a lot of media coverage in August 2017 following the change, in March of the same year, of the pharmaceutical formulation of the specialty (modification of the type and content of excipients at the initial request of the french's national agency of drug safety - ANSM). Many patients have experienced adverse effects related to fluctuating thyroid function when they were already treated and stabilized with \Levothyrox (3 million french patients with thyroid pathologies). The main idea is to find out whether the result of the processing of the data exchanged on the \Doctissimo forum can be used in pharmacovigilance and whether it would have allowed better reactivity from the  public authorities and responsible laboratories.

The work is in this paper is two-fold. First, using NLP-based techniques to extract relevant indices for potential safety signal, we attempt to underpin potential signals variations patterns during the period of the scandal. For this, we perform several analysis including (1) word frequency analysis, (2) semantic similarity analysis, (3) sentiment analysis, and (4) Adverse Drug Reactions analyses. Second, we propose a new deep CNN architecture that takes as input word clouds for detecting the occurrence of an abnormal period. The proposed architecture named Word Cloud CNN (WC-CNN) trains a CNN on word clouds extracted from the
patients comments. We study the effect of different time resolutions (day,
week, month, combined) as well as different NLP pre-processing techniques on the performance of the model. 

The contributions of this work can be summarized as follows: 
\begin{itemize}
 
\item A new Deep Learning approach based on the extraction of word clouds at different time resolutions for the early detection of pharmaceutical safety signals.
\item First approach to use word clouds as input of CNN for classification. 
\item Proposition of NLP-based indicators extracted from the patients reviews in medical forums for the early detection of pharmaceutical safety signals.
\item First analysis of the \Levothyrox scandal in 2017 in France, using natural language processing and deep learning tools. 
\item Patients comments in French on \Levothyrox Adverse Drug Reactions following the change of formula were analyzed. 
 \item Application of the proposed approach on a real world use case. 
\end{itemize}

The remaining of this paper is organized as follows. Section \ref{sec:soa} presents a literature review on AI and NLP approaches for the analysis of medical forums in the medical field. Section  \ref{sec:methods} presents the materials and methods used in this work. Section \ref{sec:results} presents the results of the study. Finally, sections \ref{sec:discussion} and \ref{sec:conclusion} discuss and conclude the paper, respectively.





\section{State of the art}
\label{sec:soa}
 Natural Language Processing (NLP) techniques have been successfully exploited for the analysis of medical forums in the medical field for various applications. However, despite the advances in data science and social networks analysis, major challenges remain to overcome to operationalize the analysis of patient posts, and efficiently support the pharmacovigilance process \cite{bousquet2017adverse,leadverse}. 
Challenges include (1) inconsistent, unstructured and region-specific data (2) variable quality of information on social media, (3) data privacy, (4) meeting expectations of pharmacovigilance experts, (5) relevant information identification on the internet, and (6) robust and evolutive architecture. In the following, we review relevant literature on data-driven pharmacovigilance research. 

\subsection{Patients' Reviews Sentiment Analysis}

Various works have focused on the study of patients' sentiment in their reviews on medical forums. Typically the polarity (positive vs. negative) of the patient's comment is detected in an attempt to understand the patients satisfaction about various aspects of their healthcare.
For instance, \cite{jimenez2019we} studied how people express their opinions about doctors and drugs in medical forums by exploring lexicon-based  and supervised learning based sentiment analysis. Models to detect sentiment polarity (positive/negative) in reviews were trained separately on drugs reviews and doctors reviews, respectively. 
It was found that drug reviews are more difficult to classify than those about physicians. The use of an informal language was found to characterize reviews about physicians. On the other hand, a combination of informal language with specific terminology (e.g. adverse effects, drug names) with greater lexical diversity was found to characterize reviews about drugs. 
Similarly, using sentiment analysis, the approach of \cite{greaves2013use} attempted to categorize the polarity of patients' online comments regarding their hospital health care (recommend a hospital, hospital was clean, good patient treatment).

In this work, we investigate the evolution in time of patients sentiment polarity as a possible indicator for safety signal detection.

\subsection{Targeted Medical Information Retrieval }

Some works tackled targeted information retrieval using clinical messages or patients reviews. For instance, an approach for relevant clinical messages filtering was proposed in \cite{bekhuis2011using}.  
NLP was used to identify clinical phrases and keywords in the messages posted to an internet mailing list. The selected phrases and keywords were then used as search strings to identify, filter and store clinically relevant messages for further analysis. Pre-processing techniques included stopwords removal, upper to lower case conversion, removal of words less than 3 characters, selection of tokens with length greater than 5  and frequency greater than 7, analysis of the 300 most frequent bi-grams and tri-grams, and sorting the messages by the number of n-gram/keywords they contain followed by computing their occurrence per year.
The work of \cite{doing2011computer} also proposed a semi-automatic update system for new candidate terms identification in live datasets for inclusion in the open access and collaborative consumer health vocabulary. The system consisted of three main parts: a Web crawler and an HTML parser, a candidate term filter that utilizes NLP techniques such as term recognition methods, and a human review interface. 

In this work, we perform a depth frequency analysis of words and bi-grams in the patients reviews. We investigate if the evolution in time of this frequency can be indicative of a safety signal. 

\subsection{ADE Identification and Extraction }

Adverse Drug Events identification and extraction from social media data has also gained attention from the pharmacovigilance research community.  Example approaches include the study drug non-compliance or use misbehavior in health online forums  using supervised classification \cite{bigeard2019detection}.  The proposed approach employed tokenization, POS-tagging, and Lemmatisation as pre-processing techniques, and NaiveBayes, Random Forest and Simple Logistic as classification methods. 
A manual analysis of the messages content has revealed that  the detected misbehavior in relation to non-compliance constitutes 28\% under-use, 27\% over-use and 6\% misuse. 

A vast amount of work have also focused on Adverse Drug Reaction (ADR) identification and extraction from social media data. 
For a survey on ADR identification and extraction in social media, the reader is referred to \cite{lardon2015adverse}. 
Existing approaches can be classified into two categories: (1) Traditional Approaches and (2) Deep Learning Approaches. 

\subsubsection{Traditional Approaches}
Existing works in this area include \cite{arnoux2019adverse} who proposed a standardized protocol for the evaluation of an NLP-based software for the extraction of adverse drug reactions (ADR) from health forums messages. ADR information extraction was performed by extracting the relation between the drug and adverse events entities, and then tested against a gold standard (manually made by two persons experienced in medical terminology). 
The approach of \cite{el2019adverse} entailed a weighted online recurrent extreme learning machine for the extraction of ADR mentions. Features were obtained by  concatenating character-level (obtained using a modified online recurrent extreme learning machine) and word-level embeddings (obtained from a pre-trained model).
An F-score of 87.5\% was obtained with this method.
Similarly, \cite{bollegala2018causality} proposed a method for ADR detection from social media based on SVM and skip-gram lexical patterns. 
Methods for filtering disorder terms that do not correspond to adverse events were also exploited in the literature.  The approach of \cite{abdellaoui2017filtering} exploited a distance-based approach where a distance (as number of words) between the drug term and the disorder or symptom term in the post was computed. An analysis of a corpus of drug-disorder pairs from 5 French forums using Gaussian mixture models and an expectation-maximization (EM) algorithm was performed. 
The results show that distance between terms can be used for identifying false positives, thereby improving ADR detection in social media.

\subsubsection{Deep Learning Approaches}
Recently, deep learning has attracted researchers to propose approaches for ADR detection and extraction from social media \cite{lee2021prediction}. 
Among the used architectures, we can find Recurrent Neural Network (RNN)  \cite{cocos2017deep}  approaches. For instance, the RNN-based approach of \cite{cocos2017deep}  labels words in an input sequence with ADR membership tags. Word-embedding vectors were used as input features.
Bi-LSTM were also proposed by \cite{fan2020adverse} for the detection and identification of professionally unreported drug side effects. The approach made use of Bidirectional Encoder Representations from Transformers (BERT) sentence embeddings  which outperformed standard deep learning architectures.

Despite the success of deep learning in ADE identification and detection, it was not explored for the task of the detection of potential safety signals. Moreover, non of the existing approaches explored the use of word cloud images as inputs to deep CNN architecture. 

\subsection{Thyroid Hormone Replacement Therapy}
Very few works have tackled the data analysis concerning thyroid hormone replacement therapy. The work of  \cite{park2018identification} used NLP to identify the themes of patient medication concerns regarding thyroid hormone replacement therapy in a dataset collected from WebMD in the United States. They used multiple regression analyses  to examine the predictability of the primary medication concerns on patient treatment satisfaction.
Their study has found six distinctive themes of patient medication concerns related to Levothyroxine treatment and that treatment satisfaction on levothyroxine was highly dependent on the primary medication concerns of the patient. As Pre-processing techniques, the approach applied stopwords removal, tokenization, stemming, and words frequency. Latent Dirichlet allocation was used to detect the topics of concerns.

Compared to the above reviewed approaches, our work falls in the category of safety signal detection in pharmacovigilance. We analyse the patients comments to extract possible indicators of a safety signal. We propose a CNN deep architecture that takes as input word cloud images extracted from patients' reviews. We explore various NLP pre-processing techniques and their effect on the performance of the deep model. The proposed approach is holistic in the sense that we look at the global behaviour of the patients' posts and its evolution in time. 

\section{Materials and Methods}
\label{sec:methods}

\begin{figure}
    \centering
    \includegraphics[scale=0.4]{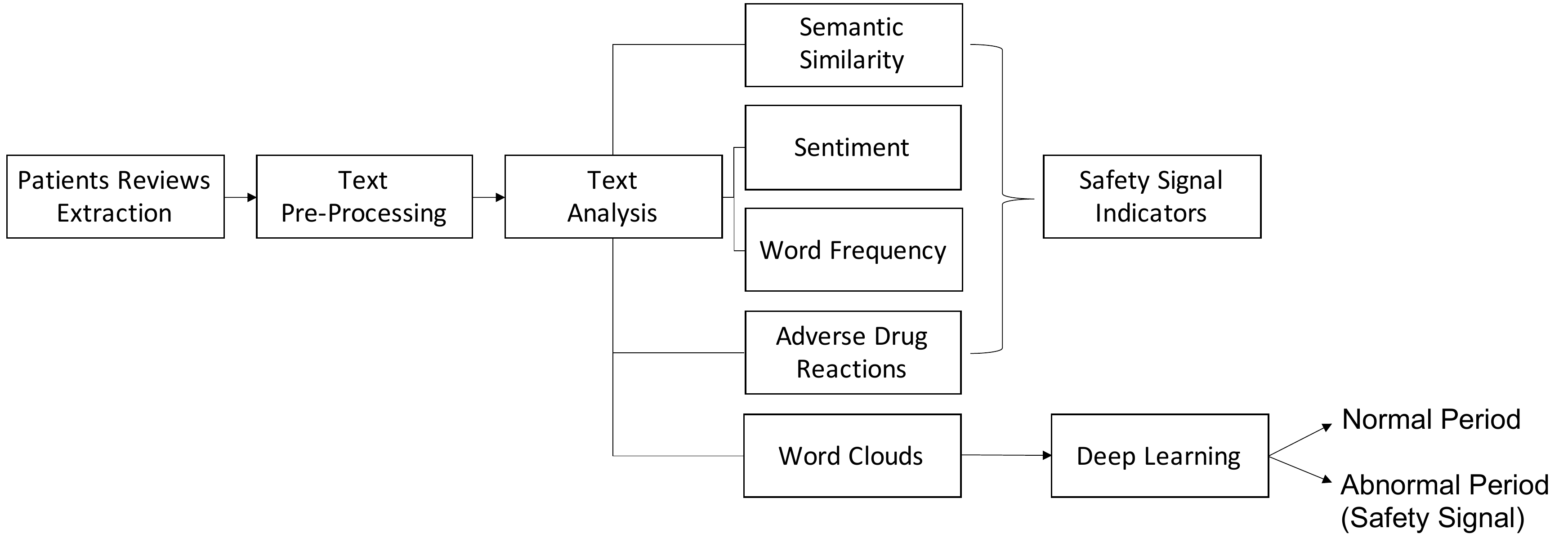}
    \caption{Workflow of the proposed approach.}
    \label{fig:workflow}
\end{figure}

In this section, we present the workflow of the proposed work (cf. Figure \ref{fig:workflow}). 
The proposed approach is composed of five steps. First in a data extraction step, a data corpus of patients reviews from an endocrinology forum is collected. NLP text pre-processing techniques are then applied to standardize the data and make it suitable for further analysis. Following, various NLP-based data analysis techniques are explored for detecting potential safety signals. Finally, we propose the Word Cloud CNN (WC-CNN) deep architecture for classifying a period as normal or abnormal. An abnormal period indicates the occurrence of a safety signal which requires further investigation.


\subsection{Data Extraction}
The data corpus was collected from the French health forum \Doctissimo \footnote{http://forum.doctissimo.fr} using a web scraping algorithm.
\Doctissimo was chosen since it is the most used health forum  in France by drug consuming patients (ranking first with $61\%$ of users). Other sources of information could have been chosen such as Twitter which is the most used website in the world by drug consuming patients. Twitter brings together 52\% of these patients against $27\%$ for all discussion forums combined. However, access to Twitter data is chargeable, which is why \Doctissimo has been selected \cite{digimind}.

The extraction was performed on the ``Thyro\"{i}de et Probl\`emes Endocriniens (Thyroid and Endocrine Problems)''  sub-forum with the keyword ``levothyrox''. The choice of extracting information from this forum using the particular relevant keyword ``levothyrox'' was to limit the amount of extracted data. Indeed, during data extraction, \Doctissimo blocks the scraping task when reaching a limit of $8,000$ extracted discussion threads, since it detects an automatic machine activity.

We collected the messages written between years $2000$ and $2020$. This resulted in a total of $110,260$  comments written by a total of $7650$ subjects. For each of the comments, we extract the date, pseudo of the person who wrote the comment, the comment's text, and  URL link. 

\begin{figure}
    \centering
    \includegraphics[scale=0.3]{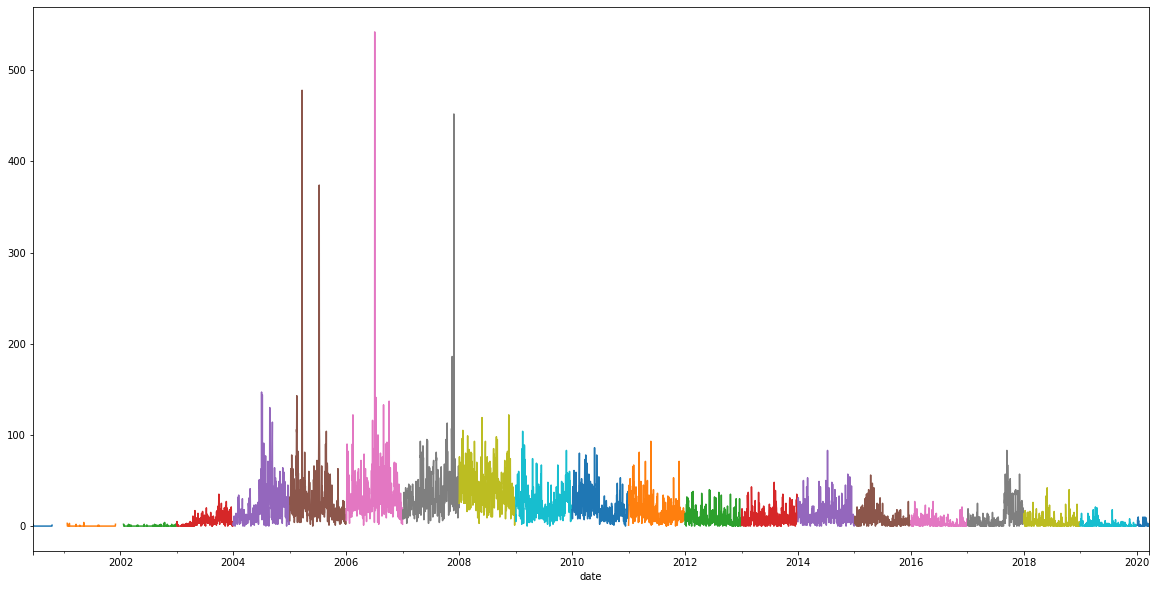}
    \caption{Number of comments posted on the \Doctissimo forum from year 2000 to 2020.}
    \label{fig:comments-fn-time}
\end{figure}

Figure \ref{fig:comments-fn-time} shows a plot of the number of comments posted on the forum between years 2000 and 2020. The figure shows that the forum was not used a lot by patients between the years 2000 and 2003. An increase of comments on the forum is event starting from year 2003. Between the years 2004 and 2012, we can see that the forum is more frequented between the years 2004 and 2012 than after 2012. It is difficult to explain the causes of this decrease in attendance except for the hypothesis that some of \Doctissimo users have migrated to other communication channels such as general social networks. Between 2012 and 2016, more comments were observed than between 2016 and 2020 but less important than before 2012. The \Levothyrox affair having taken place in 2017, it was decided to restrict the working database to the period of 2016-2020 during which the occurrence of comments is stable.

\subsection{Text Pre-processing}
Following the data extraction phase, a set of pre-processing techniques were performed on the extracted comments in order to standradize the data for further analysis and clean the dataset from noisy information that might affect the accuracy of the NLP-based analysis negatively. These include: 

\begin{enumerate}
\item \textit{\textbf{Text cleaning}}: removal of apostrophes, accents, images, emoticons, particular tags (br, span, table, strong, div, etc.), special characters (i.e. any character other than letters from A to Z and numbers from 0 to 9), digits or numbers contained in words, isolated characters, tabulations and line breaks, digits and numbers except dates, multiple spaces, unwanted patterns (those that have been identified in the database are ``\# 034'' and ``\# 039''), inserted links (identified  using the "http" or "https" pattern), and animated images (identifiable by their link ending with the pattern ``.gif.'').
Additionally  rows where at least one cell is empty were also removed from the dataset. These comments can be identified by the pattern NaN (Not a Number) or NaT (Not a Timstamp);
\item \textbf{\textit{Uppercase to lowercase conversion}}; 
\item \textit{\textbf{Spelling correction}}: correction of the spelling of the most frequent words in the database. 
First  word clouds by month, week and day visualization was performed. This allowed to visually identify the different used spellings in the database.
A word cloud allows to visually represent the most frequent words in the corpus. It represents them in different sizes according to their frequency of occurrence. Thus, the bigger the word, the more it is used in the corresponding part of the dataset (month, week, day).
Then, a dictionary was manually created with the most frequent words and their synonyms. For example the synonyms of ``levothyrox'' were identified to be ``levo'', ``levothyro'', and  ``levotyrox''. All occurrences of these words were replaced by the common synonym  ``levothyrox''. 

\item \textbf{\textit{Irrelevant words exclusion}}: a list of irrelevant words were identified and removed from the dataset. These words were qualified as undesirable/irrelevant because they are very recurrent in the comments and are neutral. They do not add any added value in terms of signal detection. These terms have  been identified manually in the text. Examples of such words include ``actuellement (actually)'', ``bonsoir (good evening)'', ``bisous (kisses)''. 

\item \textbf{\textit{Stopwords removal}}: stopwords are words so frequent that they bring a lot of  noise to the analysis. A word is qualified as empty when it is not discriminating and does not distinguish the comments from one another. A list of stopwords is created by crossing several lists freely accessible on the internet and by analyzing the database and the word clouds. The most frequent stopwords in French are ``le'', ``la", ``les'' (the), ``de'', ``du'' (of), ``ce'' (this), etc. 

\item \textbf{\textit{Lemmatization}}: lemmatization consists of reducing words to their common lemma to decrease spelling variations between words that have a similar meaning. As an example, the lemmatization transforms the words ``petit'', ``petite'', ``petits'', ``petites'' into their common lemma which is ``petit'' (small). We use Spacy library\footnote{https://spacy.io} to perform the lemmatization of the words in the corpus. 


\item \textbf{\textbf{Short comments removal}}: since a sentence has at least three words (a subject, a verb and a complement), comments that are not sentences are deleted to avoid spurious comments.

\item \textit{\textbf{Duplicate comments removal}}: sometimes users post the same message more than once. Consequently, duplicate comments were removed from the dataset.

\end{enumerate}
\subsection{Data Analysis Approach}
Various NLP analysis techniques were performed on the dataset. The goal is to investigate via various means if there is any detectable difference between the normal and the abnormal period. The abnormal period is defined to be the period where the \Levothyrox scandal happened in France, that is during the months of July and August. 
The following analyses were performed: (1) words and n-grams frequency analysis, (2) semantic similarity analysis, (3) sentiment analysis. In case a difference was detected, these analyses methods can be used as indicators of a potential safety signal and can be used as part of a pharmacovigilance surveillance system. 
\paragraph{\textbf{Words and N-grams Frequency Analysis}}
The first  explored avenue for a safety signal indicator is a frequency analysis of words in the patients' comments. The frequency analysis is carried out on single words, as well as on words sequences referred to in NLP research as n-grams \cite{sidorov2014syntactic}. 
N-grams are contiguous sequences of N elements in a sentence. N can be any positive integer. Often, N does not exceed 3 because it is rare to frequently see more than 3 adjacent words in different sentences. In this work,  bi-grams ($N = 2$) are used to know the most frequent word associations according to the different periods of the study. The purpose is to understand what are the most frequent words or words sequences occurring in the corpus, and whether there is a difference of occurring words during the different periods of the analysis. 

A correlation analysis is also performed between the yearly occurrence of the most frequent words, as well as between the most frequent bi-grams . The purpose of this analysis is to investigate whether the words or bi-grams significant to the studied use case are more correlated with each other than the others.

\paragraph{\textbf{Semantic Similarity Analysis}}

Semantic similarity measure is a metric that allows to  determine the similarity between various terms such as words, sentences, documents, concepts or instances. It allows  to find the degree of relevance between items that  are conceptually similar but not necessarily lexicographically similar \cite{ali2018semantic} \cite{harispe2015semantic}. 

In order to compute the semantic similarity of two words, first the words should be converted into numerical vectors, a process which is referred to in NLP research as word embedding. The semantic similarity is then computed using a distance measure. The most commonly used distance measure is the cosine similarity.  

We employ the Fasttext algorithm \cite{joulin2016fasttext} for learning word embeddings and computing the semantic similarity. 
FastText supports both Continuous Bag of Words and Skip-Gram models which are the most commonly used  model architectures for learning word embeddings. We implement the skip-gram model to learn vector representation of relevant words from our corpus.
The following parameters were used. 
The size of the embedding vector is set to 60. 
The window size is set to 20.
The minimum word number is set to 3. The down-sampling ratio is set to $1e^-2$. The number of iterations of 2000 is used. 

Our goal from this analysis is to investigate whether learning the embeddings from the data corresponding to each year, would result in a higher semantic similarity between the words relevant to the studied use case. For this, we train 5 word representation models, trained on the  data samples from the periods (1) years 2016 to 2020, (2) year 2016, (3) year 2017, (4) year 2018, (5) year 2019, and (6) year 2020. We then compute the semantic similarity between the identified relevant word to the \Levothyrox scandal.

\paragraph{\textbf{Sentiment Analysis}}
 Also using the FastText library, we perform sentiment analysis of the patient's comments. We train a classifier to detect the polarity of a comment (positive/negative). Our aim is to study the evolution of the patients' sentiment in function of time and examine if there is an evident increase of the negative sentiments expressed in the patients' comments during the period of the scandal. 
 
 The training of the sentiment detection  algorithm is carried out on a database of French tweets already labeled positive or negative \cite{frenchTwitterData}. Since French social media datasets are scarce, this dataset was generated by translating an existing English tweets dataset to French. 
 The polarity of each forum post is then predicted by running the trained algorithm. 
 
For the sentiment detection model to be as efficient as possible, it must be trained on a database in the same language and on a subject most similar to the database to be labeled. Ideally, the tweets would be in French and touch on medical topics. However, no database meeting these two criteria was found. It was therefore decided to focus on the general French tweets database.

\paragraph{\textbf{Adverse Drug Reactions Analysis}}
We study the evolution of Adverse Drug Reactions occurrence in the patients comments. 
For this we develop a method for the detection of Adverse Drug Reactions. The method is based on regular expressions. Regular expressions (regex or regexp) \cite{regex} are sequences of characters specifying a search pattern. First we define a list of possible Adverse Drug Reactions related to hypothyriodism. 
Then using a string-searching algorithm, we detect and count the occurrence of the defined Adverse Drug Reactions in the patients comments.


\subsection{Deep Learning Approach}
The safety signal prediction problematic is formulated as a classification problem with two classes: normal and abnormal. A normal period is defined to be a period where there was no event triggering an increase  in the frequency of adverse drug reactions normally reported by patients. In other words, it refers to a period where the behaviour of patients is considered as normal. The abnormal period on the other hand, is defined to be a period where an important event has happened and triggered an alteration of the reporting behavior of patients on the medical forum. Such behavior is an indication that a safety signal should be flagged, to incite attention and consequently a verification or other type of action. 

\begin{figure}
    \centering
    \includegraphics[scale=0.3]{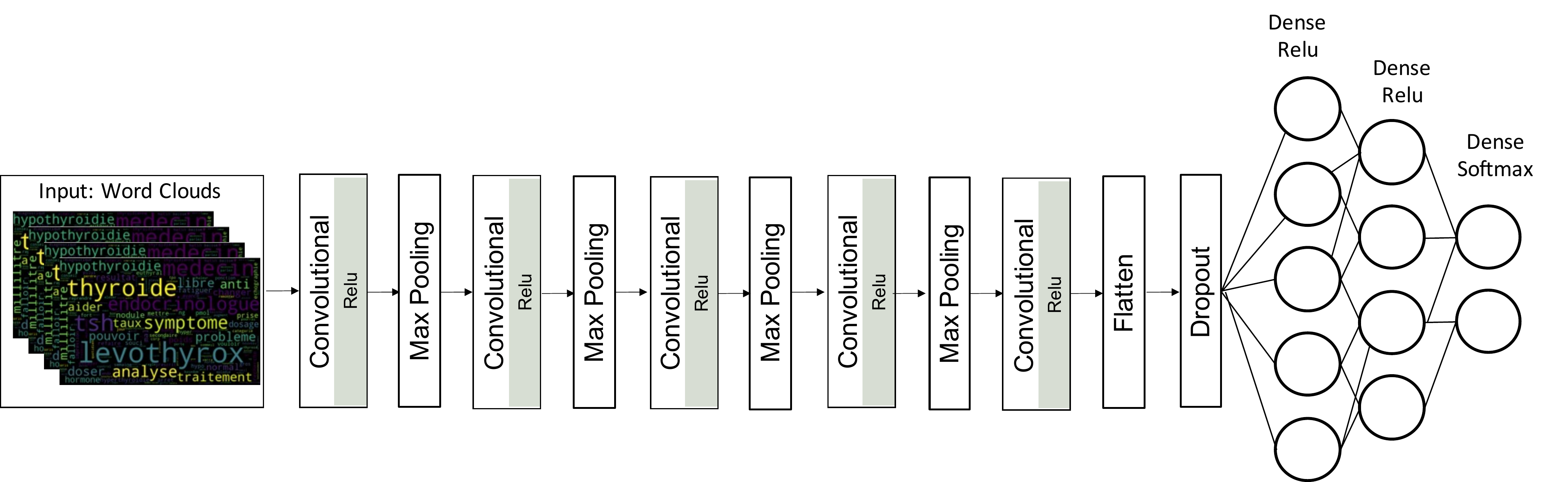}
    \caption{Proposed CNN architecture.}
    \label{fig:cnn}
\end{figure}

The proposed deep neural network, referred to as Word Cloud CNN (WC-CNN) is an architecture allowing to learn high level features from word clouds extracted from the patients' reviews at different time resolutions. Word clouds being visualizations of the most frequent words in a given text with their size reflecting their importance and frequency, represent an avenue to explore Convolutional Neural Networks for classification. The output of the WC-CNN is a binary label (0 for abnormal, and 1 for normal period). 
 Figure \ref{fig:cnn} presents an overview of the proposed architecture and the following details the word clouds extraction and the proposed architecture. 

\subsubsection{Word Clouds Extraction}
A word cloud is one of the most popular data visualization techniques used to represent textual data. It allows to visually represent the most frequent words where the size of a word indicates its frequency or importance in the text. 
We generate word clouds of size $800 \times 500$ with a maximum number of words of $200$.
The maximum font size  for the largest word is set to 110. The minimum font size is set to 4. A Relative Scaling (RS) of 0.5 is used.
The Font Size (FS) is computed in function of the frequency as follows:

\begin{equation}
    FS = (RS * (\frac{frequency}{lastfreq}) + (1 - RS)) * FS
\end{equation}

\subsubsection{WC-CNN Architecture}
The word cloud features extracted from the patients' comments are fed to a Convolutional Neural Network (named: Word Cloud CNN) to extract high-level deep features. The deep architecture is composed of four two-dimensional (2D) convolutional layers each followed by a ReLU activation function and a max pooling layer, a fifth convolutional layer, a flatten layer, and a dropout layer. 

\paragraph{Network Implementation Details}

The number of convolution filters for all convolutional layers is 128. The first 3 convolutional layers have filter sizes of 6 x 6.  the fourth and fifth convoluitional layers have filter sizes of 3x3 and 2x2 respectively. A pool size of 2x2 was used for all pooling layers. 


RELU is used as activation function for all convolutional and fully connected layers. 
The output layer is a dense layer of size 2 with a Softmax activation function. 
 The proposed models are trained with the categorical cross entropy as loss function and the Adam optimizer. 
 The batch size is set to 50 samples. 
 The number of epochs for training is set to 100. 
 An early stopping is performed when the loss function stops improving after 10 epochs.
 
\section{Results}
\label{sec:results}
In this section, we present the results of the performed data analysis presented in the above section. Moreover, the results of the proposed deep learning model are also presented and discussed. 
\subsection{Data Analysis Results}

\subsubsection{Words and N-grams Frequency Analysis}
We present the results of the performed frequency analysis over the dataset. The frequency analyzes carried out are as follows:
\begin{enumerate}
    \item Word frequency analysis over the entire studied period  (2016 to 2020).
    \item Analysis of bi-grams over the entire studied period  (2016 to 2020).
    \item Analysis of the correlation between the occurrence of the most frequent words.
    \item Analysis of the correlation between the occurrence of the most frequent bi-grams.
\end{enumerate}
\paragraph{\textbf{Words Frequency Analysis}}

\begin{figure}
    \centering
    \includegraphics[scale=0.6]{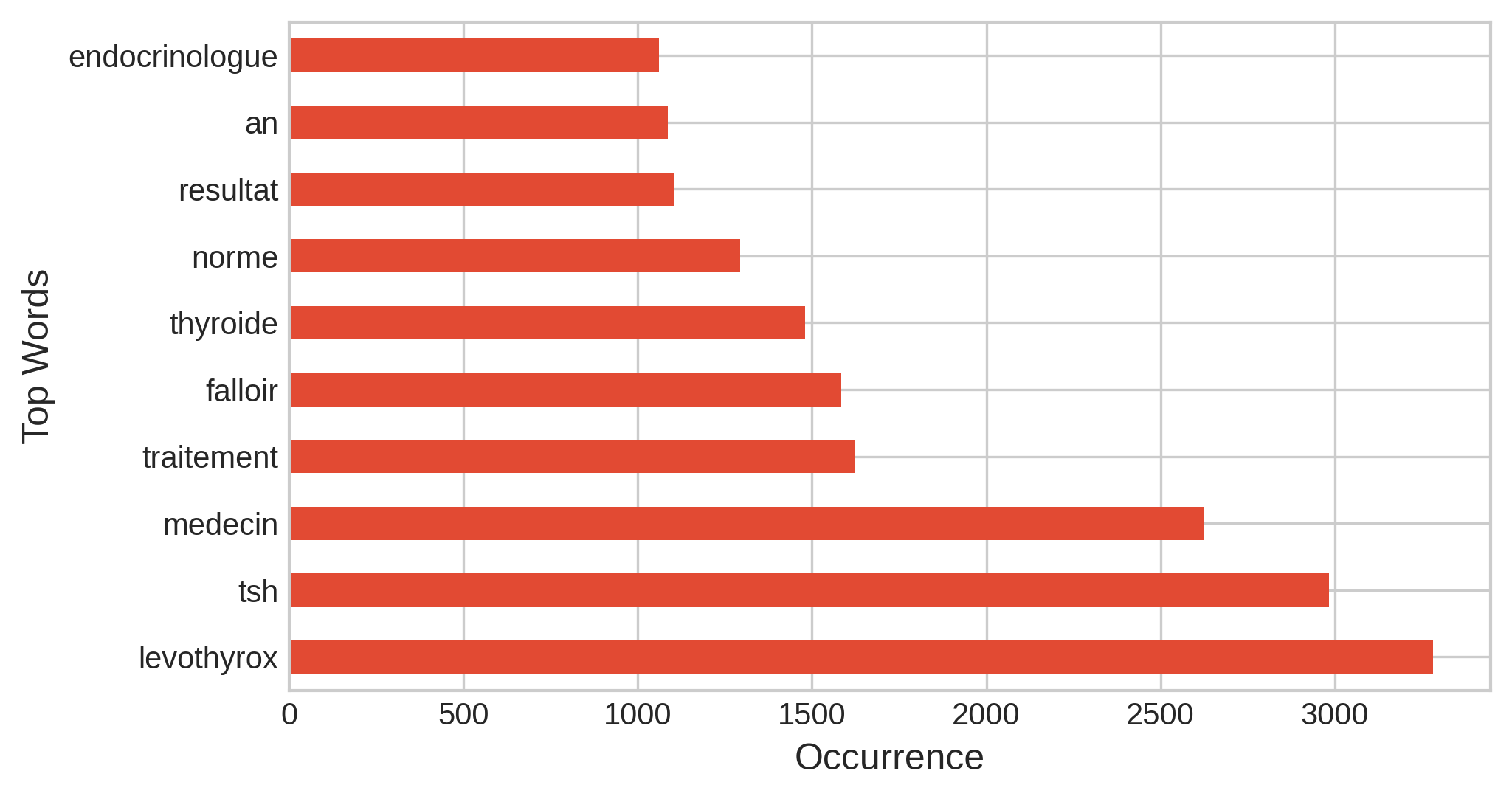}
    \caption{Top word occurrence (period 2016-2020).}
    \label{fig:top_word_occ}
\end{figure}

Figure \ref{fig:top_word_occ} shows the top 10 frequently occurring words in function of their frequency during this period. Among these words, we can find the words ``levothyrox'', ``tsh'', ``medecin (doctor)'', ``traitement (treatment)'', ``thyroide'', ``resultat (result)'', and ``endocrinologue (endocrinologist)''. This allows to validate that the topics appearing in the patients' comments the most are related to the problems concerning the medication, and treatment.   

\begin{figure}
    \centering
    \includegraphics[scale=0.7]{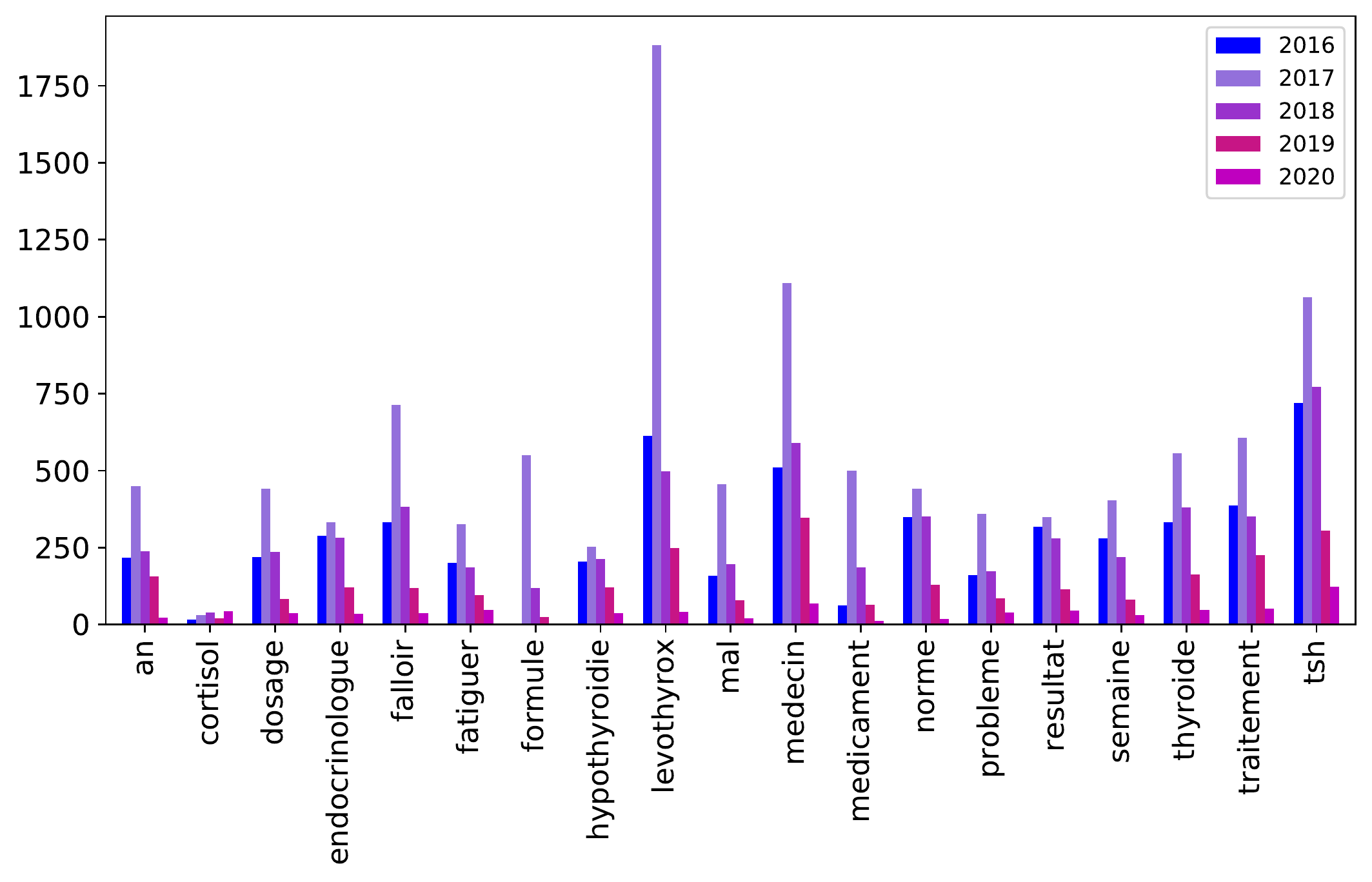}
    \caption{Top word occurrence per year (period 2016-2020).}
    \label{fig:top_word_hist}
\end{figure}

Figure \ref{fig:top_word_hist} shows the most frequently observed words over five years (2016-2019) and their frequency of occurrence in each year. From the figure, it is observed in 2017 (year of \Levothyrox scandal) that the word ``dosage'' or ``doser (dose)'' occurs $1.74$ times more than in 2016 and $3.49$ times more than the years after 2017. Similarly for the words ``fatiguer (tired)'' ($1.63$ vs. $2.63$), ``formule (formula)'' ($274$ vs. $14.22$) , ``levothyrox'' ($3.23$ vs. $5.72$), ``mal (pain)'' ($2.86$ vs. $4.00$), ``medecin (doctor)'' ($2.20$ vs. $2.45$), ``symptome (symptom)'' ($1.70$ vs. $2.12$), and ``traitement (treatment)'' ($1.53$ vs. $2.11$). 

The observation of these results clearly indicates strong user activity in connection with \Levothyrox in 2017. Many of the identified words  have an over-representation of their occurrences during 2017; starting with the  example of a ``formula'' which was only used twice in 2016 and 548 times the year after. With these results, it was indeed possible to deepen this work by using the undesirable effects and their related significant terms (e.g. ``dosage'', ``doser (dose)'', ``fatiguer (tired)'', ``mal (pain)'', ``symptome (symptom)'', ``traitement (treatment)''). 

\begin{figure}
    \centering
    \includegraphics[scale=0.7]{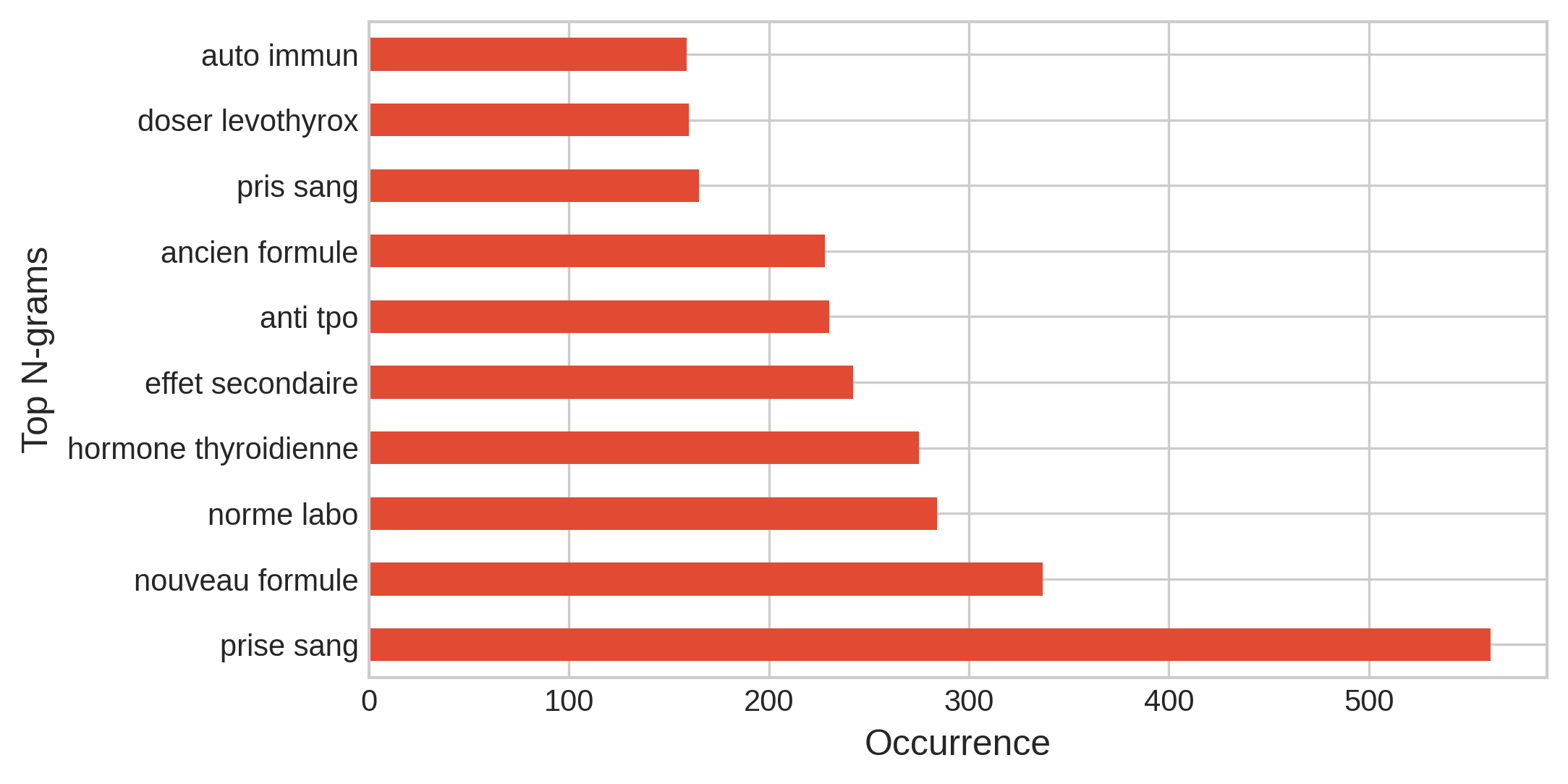}
    \caption{Top n-gram occurrence (period 2016-2020).}
    \label{fig:top_ngram_occ}
\end{figure}

\begin{figure}[!ht]
    \centering
    \includegraphics[scale=0.6]{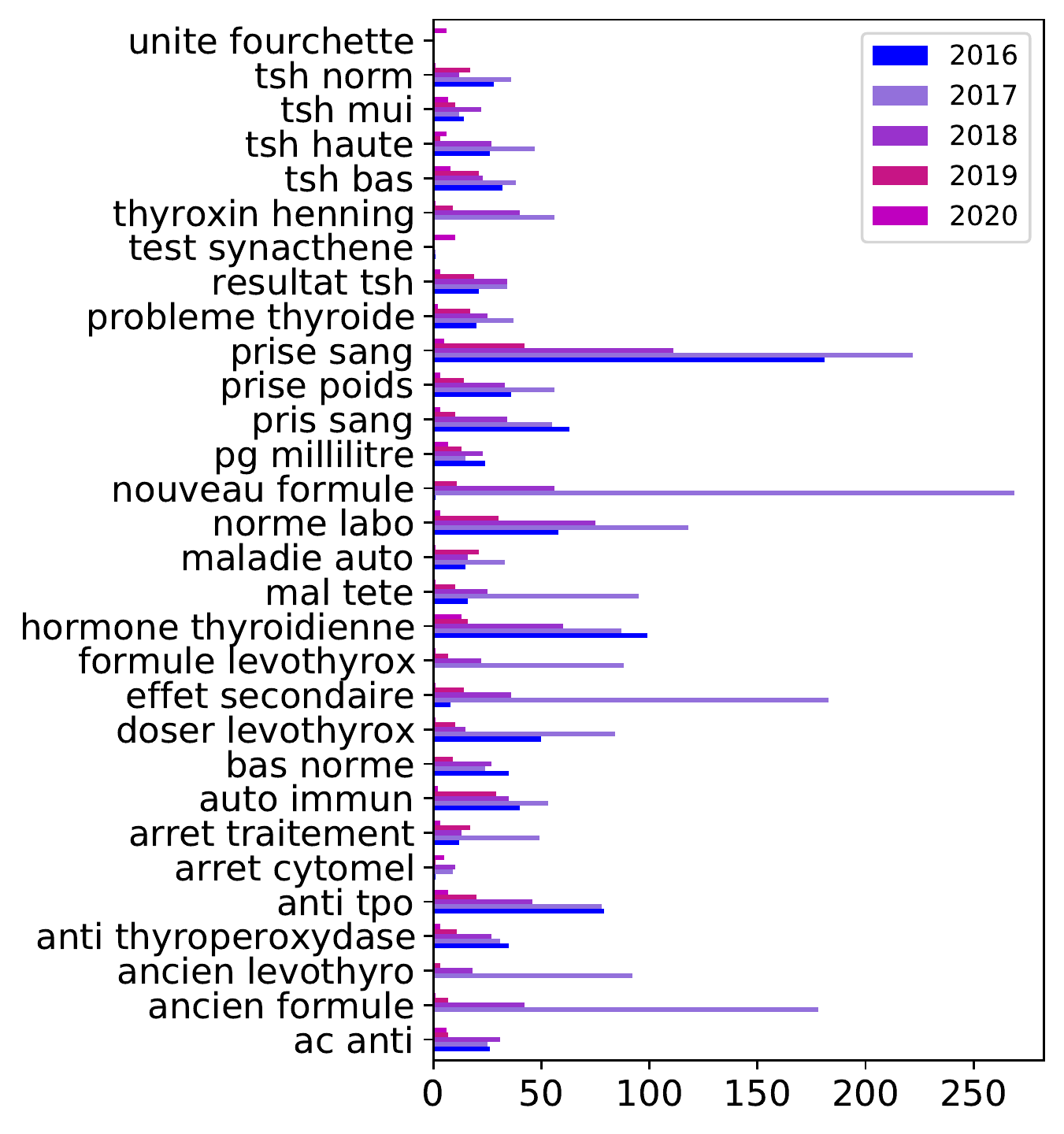}
    \caption{Top n-gram per year (period 2016-2020).}
    \label{fig:top_ngram_hist}
\end{figure}

\paragraph{\textbf{Bi-grams Frequency Analysis}}
Figure \ref{fig:top_ngram_occ}  shows the 10 most frequent bi-grams over the whole studied period. Among these we can notice that ``nouveau formule (new formula)'', ``effet secondaire (secondary effect)'', and ``ancien formule (old formula)'' are among the bi-grams appearing during this period. 
Figure \ref{fig:top_ngram_hist} shows the most frequently observed bi-grams over five years (2016-2019) and their frequency of occurrence.
The analysis made through the bi-grams is more relevant than the over the individual words occurrence presented in the previous section. The terms displayed are consistent and more in line with what is expected.
Certain bi-grams are particularly relevant to the studied scandal such as ''ancien formule (old formula)``, ''effet secondaire (side effect)``, ''formule levothyrox (levothyrox formula)``, and ''nouveau formule (new formula)``. The figure clearly shows that in the year of the scandal (2017) the number of occurrence of these bi-grams is much higher than their occurrence during the years. 

Table \ref{tab:top10_ngram_peryear} shows the 10 most frequent bi-grams in years 2016 to 2020. We can notice that even more convincingly, bi-grams that did not exist in 2016 become some of the most observed bi-grams of the year of the formula change. These include bi-grams that are directly related to the studied scandal, such as ``nouveau formule (new formula)'', ``ancien formule (old formula)'', ``ancien levothyrox (old levothyrox)'' and ``levothyrox formula (formule levothyrox)''.

\begin{table}
 \caption{The 10 most frequent bi-grams in years 2016 to 2020.} 
\tiny
  \centering
\begin{tabular}{lllll}
\toprule
\textbf{2016} & \textbf{2017} & \textbf{2018} & \textbf{2019} & \textbf{2020} \\ \midrule
prise sang                              & nouveau formule                         & prise sang                              & prise sang                              & hormone thyroidienne                    \\
hormone thyroidienne                    & prise sang                              & norme labo                              & norme labo                              & test synacthene                         \\
anti tpo                                & effet secondaire                        & hormone thyroidienne                    & auto immun                              & tsh bas                                 \\
pris sang                               & ancien formule                          & nouveau formule                         & maladie auto                            & anti tpo                                \\
norme labo                              & norme labo                              & anti tpo                                & tsh bas                                 & pg millilitre                           \\
doser levothyrox                        & mal tete                                & ancien formule                          & anti tpo                                & tsh mui                                 \\
auto immun                              & ancien levothyro                        & thyroxin henning                        & resultat tsh                            & ac anti                                 \\
prise poids                             & formule levothyrox                      & effet secondaire                        & arret traitement                        & tsh haute                               \\
anti thyroperoxydase                    & hormone thyroidienne                    & auto immun                              & probleme thyroide                       & unite fourchette                        \\
bas norme                               & doser levothyrox                        & pris sang                               & tsh norm                                & arret cytomel                           \\ \bottomrule
\end{tabular}
\label{tab:top10_ngram_peryear}
\end{table}

\paragraph{\textbf{Correlation Analysis}}
Table \ref{tab:bigram_occ_corr} displays the highest correlations  between the 2 bigrams among the top identified bi-grams during the years 2016-2020. 

The 10 strongest correlations between two bi-grams are between the identified most occurring bi-grams (cf. figure \ref{fig:top_ngram_hist}) and concern the change in formula of \Levothyrox.

\begin{figure}
    \centering
    \includegraphics[scale=0.4]{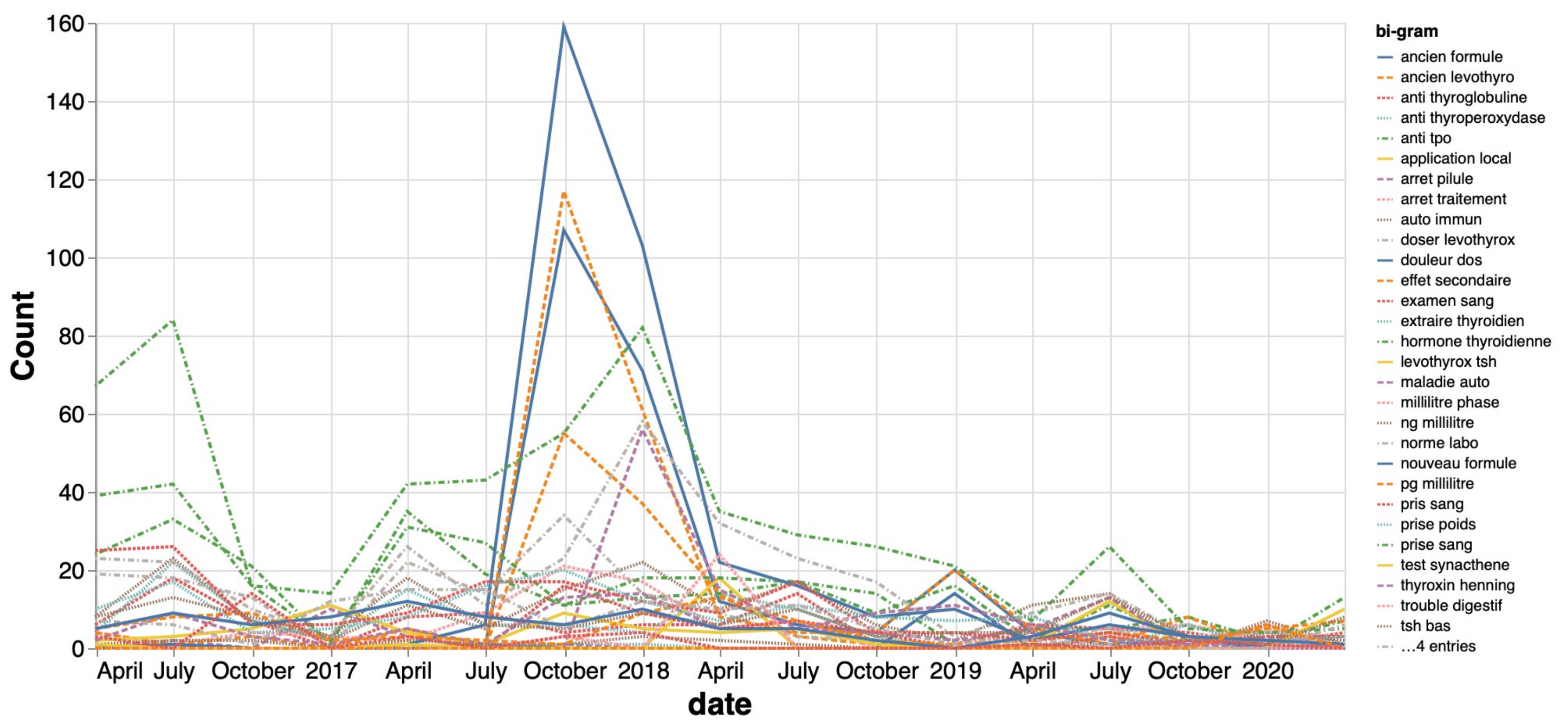}
    \caption{Histogram reflecting the frequency of appearance of "top n-gram" for the period 2016-2020.}
    \label{fig:hist_top_ngram}
\end{figure}

This is undeniable proof of what could be speculated: something new happened in 2017, to the point of dominating the trends over a four-year period. In addition, looking at a monthly breakdown, it can be seen in Figure \ref{fig:hist_top_ngram} that this phenomenon started with media announcements, in July 2017.

The study of this histogram shows that using a correlation study of bi-gram occurrences, we can perform an early detection of possible pharmaceutical safety signals. 

\begin{table}
\footnotesize
  \centering
  \caption{The highest correlations  between the 2 bigrams among the top identified bi-grams during the years 2016-2020.}
\begin{tabular}{@{}llr@{}}
\toprule
\textbf{Bi-gram 1} & \textbf{Bi-gram 2} & \multicolumn{1}{l}{\textbf{Correlation}} \\ \midrule
ancien levothyro   & nouveau formule    & 0.9999287212                             \\
ancien formule     & nouveau formule    & 0.9995480603                             \\
ancien formule     & formule levothyrox & 0.9993103979                             \\
ancien formule     & ancien levothyro   & 0.9992348913                             \\
ancien levothyro   & effet secondaire   & 0.9989443185                             \\
effet secondaire   & nouveau formule    & 0.9988471039                             \\
formule levothyrox & nouveau formule    & 0.9988416757                             \\
ancien levothyro   & formule levothyrox & 0.9982914213                             \\
ancien formule     & effet secondaire   & 0.9971094253                             \\
effet secondaire   & formule levothyrox & 0.9966794925                             \\ \bottomrule
\end{tabular}
\label{tab:bigram_occ_corr}
\end{table}

\subsubsection{Semantic Similarity Analysis}
\begin{figure}
    \centering
    \includegraphics[scale=0.5]{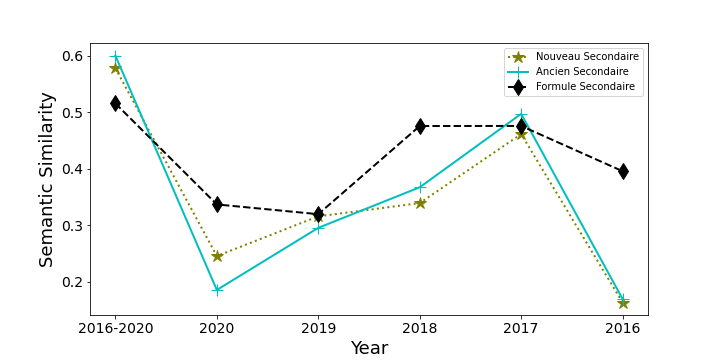}
    \caption{Semantic similarity analysis (period 2016-2020).}
    \label{fig:semantic_similarity}
\end{figure}

Figure \ref{fig:semantic_similarity} shows the evolution of the semantic similarity between relevant words to the \Levothyrox scandal. We compute and plot the semantic similarity between the words ``Nouveau (new)'' and ``Secondaire (secondary)'', ``Ancien (old)'' and ``Secondaire'', ``Formule (formula)'' and ``Secondaire''. The word representations were learned on the data corresponding to all years (2016-2020), and only the data corresponding to each year. 
The figure shows that year of the scandal (2017) shows the highest semantic similarity score with the words of interest. 
\subsubsection{Sentiment Evolution Analysis}
\begin{figure}[!ht]
    \centering
    \includegraphics[width=\textwidth]{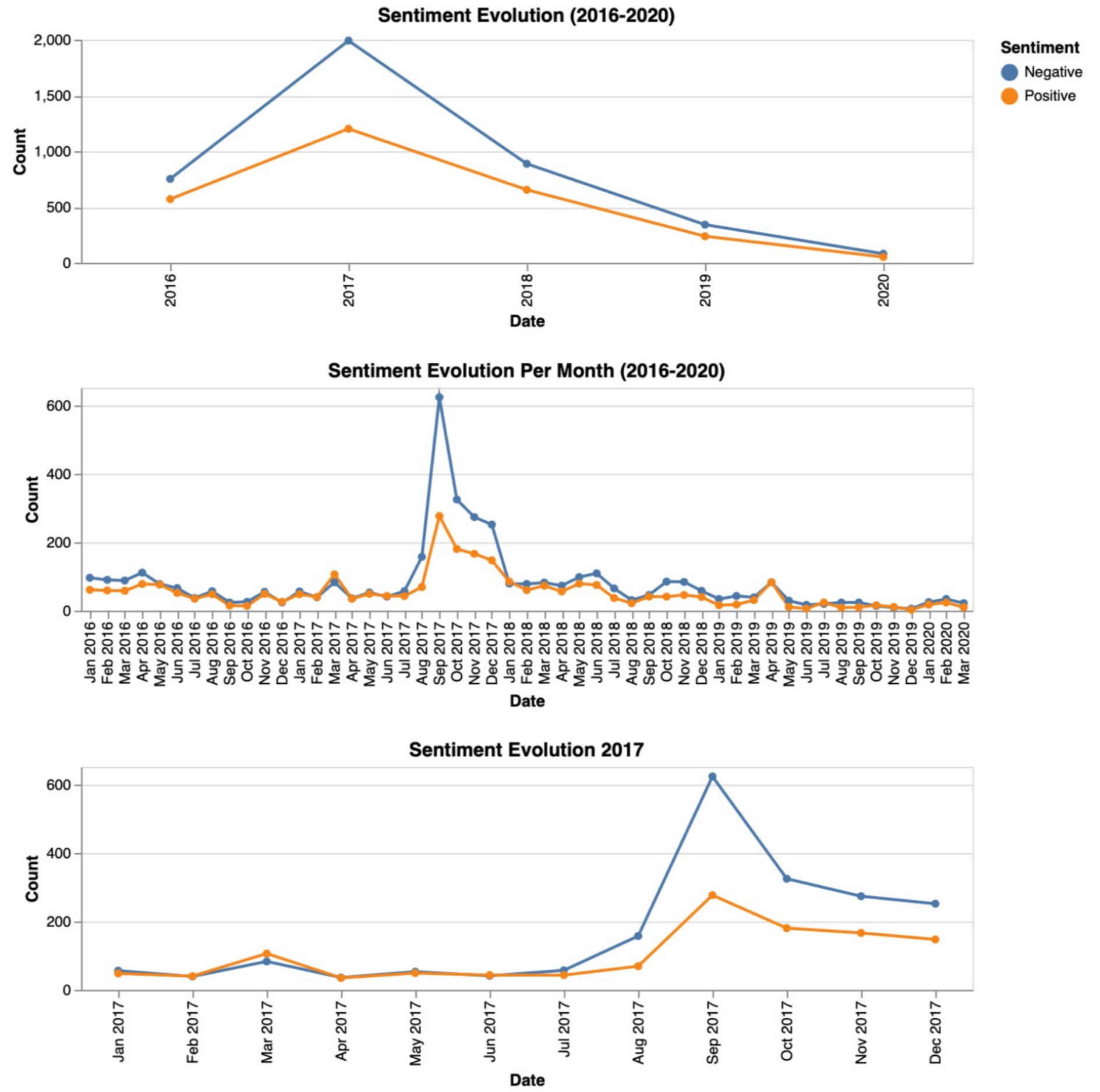}
    \caption{Evolution of positive and negative sentiments in the patients' comments per year from 2016 to 2020 (top), per month from 2016 to 2020 (middle), and per month during the year 2017.}
    \label{fig:sent_evolution}
\end{figure}

We analyze the evolution of the perceived sentiments in the user's comments. Figure \ref{fig:sent_evolution} shows the evolution of negative and positive sentiments in the patients' comments per year from 2016 to 2020 (top), per month from 2016 to 2020 (middle), and per month during the year 2017. The figure clearly shows an increase of negative sentiments in the patients' comments during the year of the scandal (2017). Zooming on the number of sentiments taking into account a monthly resolution confirms that patients comments were negative during the period of Levothyrox formula change.

\subsubsection{Adverse Drug Reactions Analysis}
\begin{figure}
    \centering
    \includegraphics[scale=0.7]{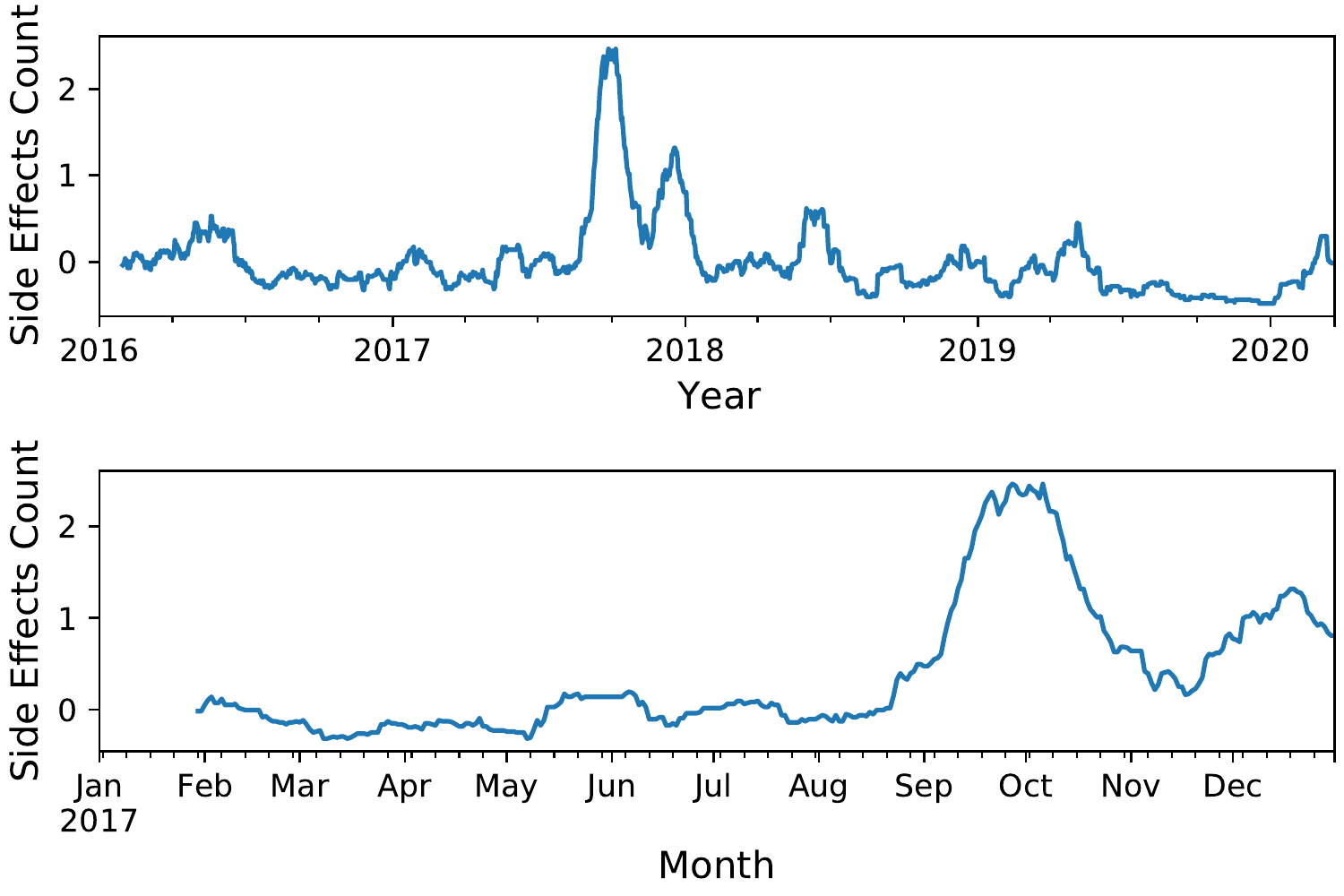}
    \caption{Adverse Drug Reactions occurrence in the patients comments between 2016 and 2020 (upper plot) and during the year 2017 (lower plot).}
    \label{fig:se_2016_2020}
\end{figure}

\begin{figure}
    \centering
    \includegraphics[scale=0.3]{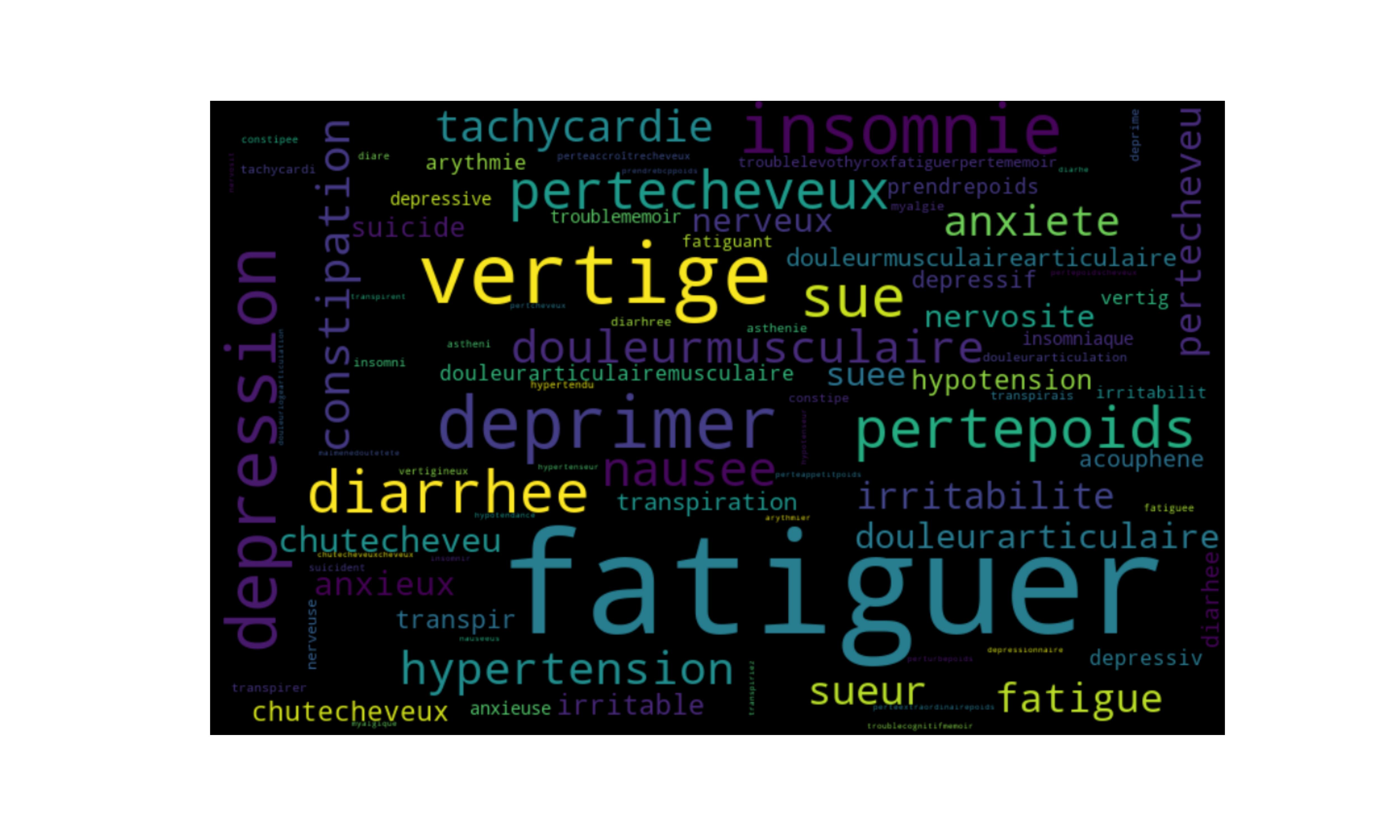}
    \caption{Visualization of the most common Adverse Drug Reactions detected from the corpus (period 2016-2020).}
    \label{fig:mcse}
\end{figure}

\begin{figure}
    \centering
    \includegraphics[scale=0.7]{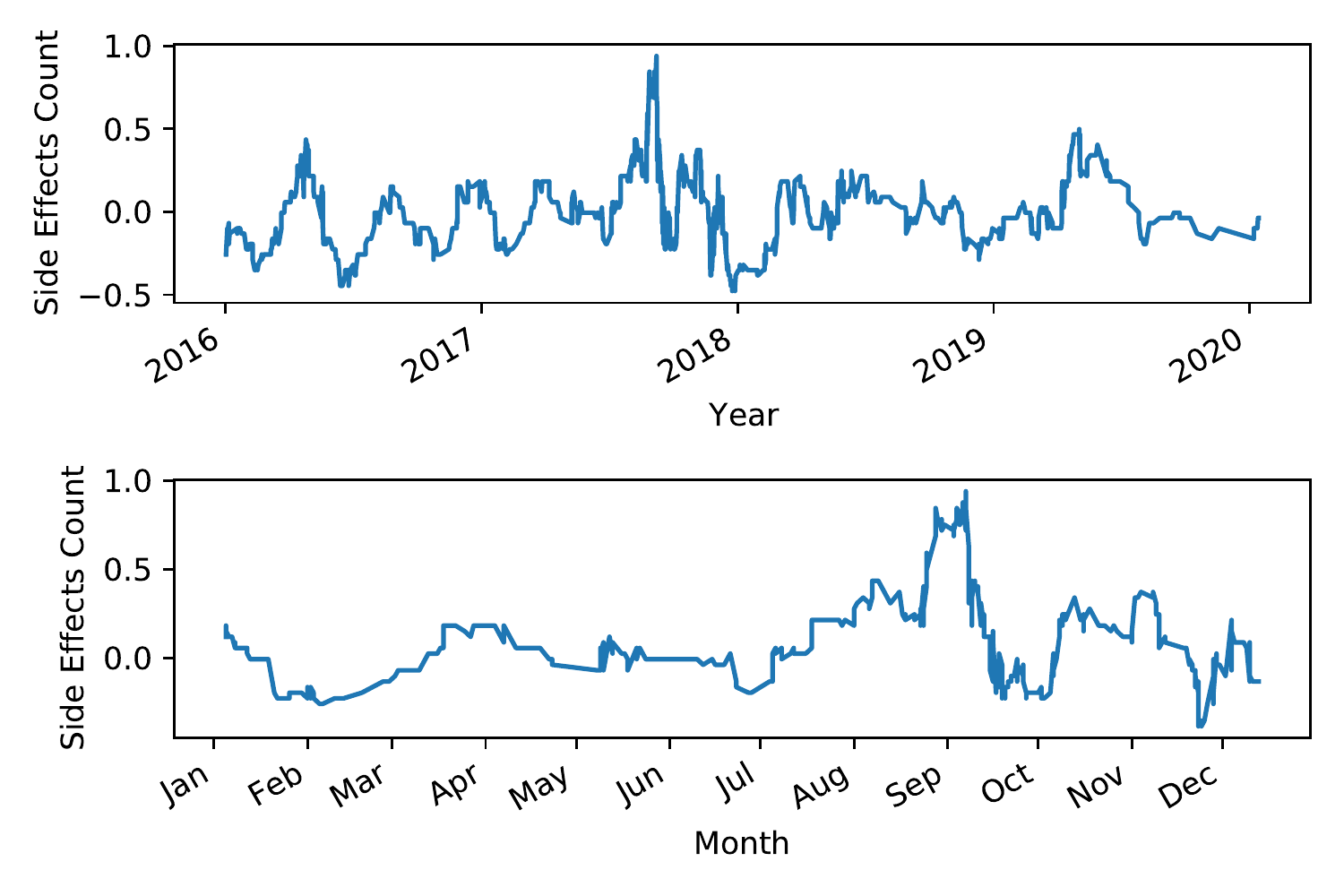}
    \caption{Most common Adverse Drug Reactions occurrence in the patients comments between 2016 and 2020 (upper plot) and during the year 2017 (lower plot).}
    \label{fig:mcse_2016_2020}
\end{figure}

\begin{figure}
    \centering
    \includegraphics[scale=0.5]{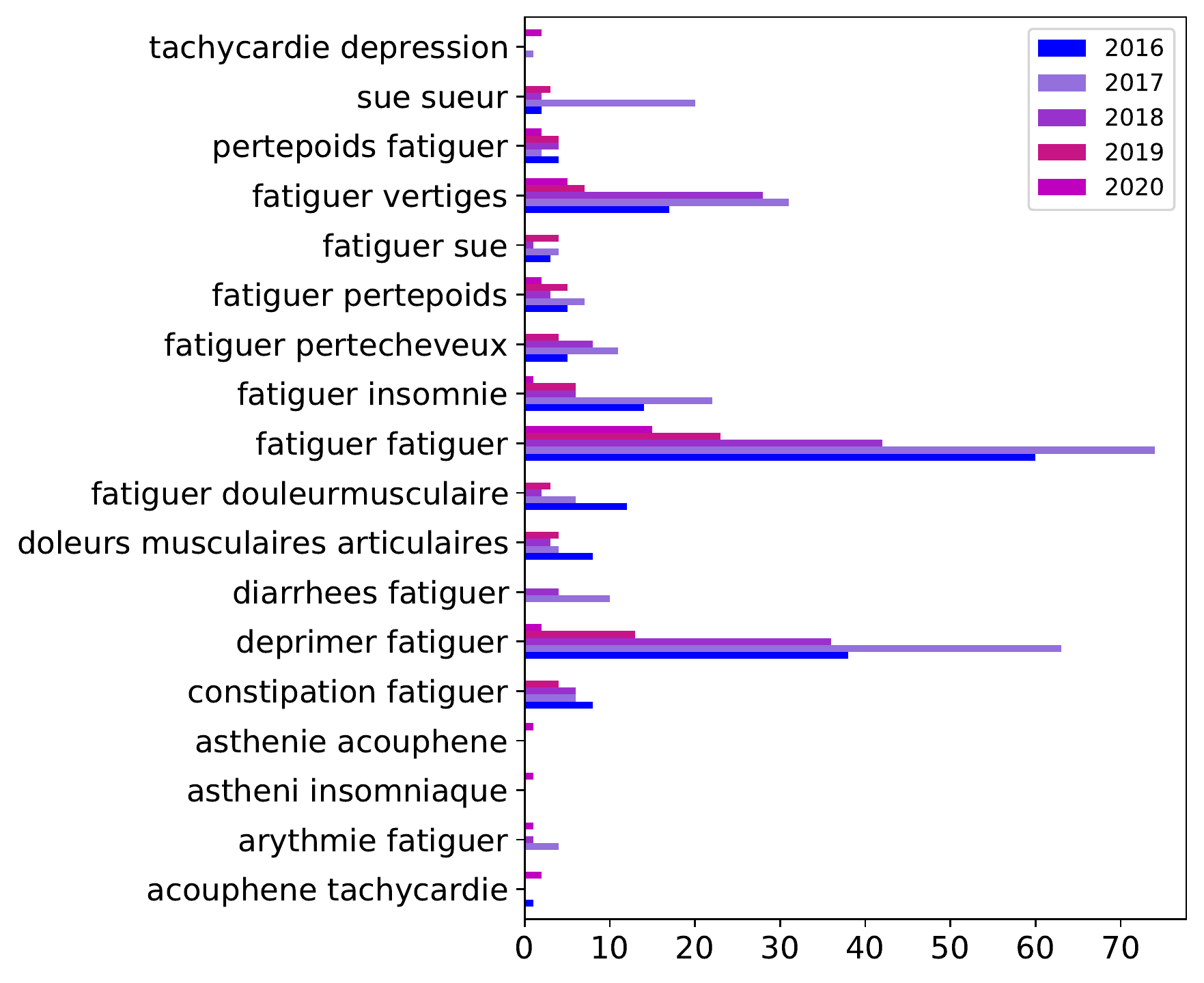}
     \caption{Top n-gram Adverse Drug Reactions (period 2016-2020)}
    \label{fig:top_ngram_se}
\end{figure}
Figure \ref{fig:se_2016_2020} shows the Adverse Drug Reactions occurrence count in the patients comments between 2016 and 2020 (upper plot) and during the year 2017 (lower plot). 
The plot of the Adverse Drug Reactions count using a yearly resolution shows that there are some periods that witness an increase of the Adverse Drug Reactions occurrence in the patients reviews. This increase might be due to several factors depending on the event that triggered the patients to complain about the Adverse Drug Reactions that they are experiencing. One notable peak happened at the end of year 2017 which coincides with the \Levothyrox scandal. 
Zooming on the Adverse Drug Reactions occurrence during this year. The plot shows that the number of Adverse Drug Reactions started to increase from the middle of August and reached it peak beginning of September, lasting till October where we start to see a decrease of the count of Adverse Drug Reactions. 

We detect the most common Adverse Drug Reactions and perform the same analysis as above. Figure \ref{fig:mcse} shows the word cloud of the the most common Adverse Drug Reactions detected from the corpus. Figure \ref{fig:top_ngram_se} also shows the top n-gram Adverse Drug Reactions (period 2016-2020).
Figure \ref{fig:mcse_2016_2020} shows the plots of the counts of the most common Adverse Drug Reactions. The plots show the same trend detected in figure \ref{fig:se_2016_2020}.

\subsection{CNN Performance Evaluation}
We explored different combinations of pre-processing techniques and time resolutions to determine which combination resulted in the highest predictive accuracy for abnormal period detection.

In order to test the performance of the neural network, the word clouds were labeled normal or abnormal according to three periods of abnormality:

\begin{enumerate}
    \item July to December 2017: information is relayed in the media.
    \item May 2017 to February 2018: we increased the abnormal period duration by 2 months from each side (before and after the information was relayed in the media).
    \item March 2017 to April 2018: 2 months period around the second period.
\end{enumerate}

In order to obtain balanced datasets, the normal period is selected taking into account the number of data samples in the abnormal period. We select the same number of abnormal data sample from the period before and after the abnormal period. 

For the above defined three abnormal periods, four different word cloud pre-processing techniques are applied to test whether the variations in the pre-processing have an effect on the prediction performance. This would inform the best combination of pre-processing for the deep learning model to learn to differentiate the two periods. 
Word clouds are produced at three different frequencies: daily, weekly and monthly, in order to know the most relevant data to use to obtain the best prediction results. The advantage of defining three periods of abnormality is to identify whether an early detection of the problem related to the change in formula of \Levothyrox is possible.

This early detection can be confirmed if:
\begin{itemize}
\item The performance of the neural network is good over the period July to December 2017. This means that there is indeed a difference between the clouds of words labeled normal and abnormal.
\item The results observed over the period May 2017 to February 2018 and/or March 2017 to April 2018 are as good as over the period July to December 2017. This confirms that the word clouds of two or three periods of abnormality are similar (as the three abnormal periods include the period July to December 2017)
\end{itemize}

If the two previous conditions are met, it can be concluded that a difference with the period of normality is identifiable and that early detection is possible.

\subsubsection{Time Resolution Effect}
For the July to December 2017 abnormality period, the highest prediction is obtained via monthly word clouds (accuracy 0.667), followed by weekly word clouds with accuracy 0.611. The same conclusions are drawn for the period of abnormality May 2017 to February 2018 with three very good results of 0.75 and 0.65 for monthly word clouds and 0.625 for weekly word clouds. For the period March 2017 to April 2018, the results are a little worse but remain acceptable. During tests with other neural networks, the period March 2017 to April 2018 frequently shows results a little below the other two. When all the word clouds are combined, the results are worse and almost identical to those observed with daily word clouds (still around 0.5). This observation is explained by the very large number of daily word clouds compared to weekly and monthly which greatly reduce the performance of prediction.

\subsubsection{Data Pre-processing Effect}

\begin{enumerate}
\item Remove comments of less than three words as a sentence contains at least one subject, one verb, and one complement.
\item Removal of "stopwords".
\item Lemmatization of comments.
\item Improvements in lemmatization by creating several lists to correct the spelling of words observed in word clouds.
\item Creation of a list called $"wordstodelete"$, intended to clean the word clouds of parasitic words unrelated to the medical context.
\end{enumerate}

Data pre-processing has little impact on the prediction performance of neural networks. Indeed, results greater than 0.6 are observed with all the different cleaning operations. These observations are also found during tests carried out by varying the parameters of the neural networks.
Thus, it is observed that the prediction performance increases with increasing number of convolution layers and is better on weekly and monthly word clouds. It is not observable on this neural network but during the various tests carried out, it is identified that the monthly word clouds present better results than the weekly ones.
Also, the results of the periods of abnormality July to December 2017 and May 2017 to February 2018 being just as good, it is possible to conclude that an early detection of signals indicating abnormal events is identifiable from the month of May 2017. It is noted that these results are less good over the period of abnormality from March 2017 to April 2018. This poorer result can be explained because the new formula arrived on the market in March and was still little consumed by patients. Also, the end of the period of abnormality extending until April 2018, the patients exchanged less on the subject after December 2017. The word clouds are less close to those observed between July and December 2017 which is the period at during which an explosion in the frequency of comments is observed on \Doctissimo.

\begin{table}
\caption{WC-CNN performance for the detection of abnormal periods (safety signals) using different combinations of pre-processing techniques and different training periods data.}
\centering
\begin{tabular}{lllll}
\multicolumn{5}{c}{{ \textbf{Abnormal Period: July to December 2017}}}                                                                               \\
\textbf{Pre-processing} & \multicolumn{1}{c}{\textbf{Day}} & \multicolumn{1}{c}{\textbf{Week}} & \multicolumn{1}{c}{\textbf{Month}} & \multicolumn{1}{c}{\textbf{All periods}} \\
\hline
 Steps 1 2 3 4 5            &  0.557     & 0.519                                & 0.250                             &  0.572                           \\
 Steps 1 2 3 4              &  0.554     &  0.611        & 0.500                             & 0.527                                                   \\
 Steps 1 2 3 5              & 0.548                             &  0.593        &  \textbf{0.667}     & 0.493                                                   \\
 Steps 1 2 5                & 0.533                             & 0.463                                & 0.333                             & 0.527                                                   \\
 \hline
\multicolumn{5}{c}{{ \textbf{Abnormal Period: May 2017 to February 2018}}}                                                                             \\
\textbf{Pre-processing} & \multicolumn{1}{c}{\textbf{Day}} & \multicolumn{1}{c}{\textbf{Week}} & \multicolumn{1}{c}{\textbf{Month}} & \multicolumn{1}{c}{\textbf{All periods}} \\
\hline
Steps 1 2 3 4 5            & 0.531                             &  \textbf{0.625}        & \textbf{0.750}     &  0.566                           \\
Steps 1 2 3 4              & 0.547                             & 0.523                                &  0.550     & 0.537                                                   \\
Steps 1 2 3 5              & 0.494                             & 0.455                                & 0.400                             & 0.519                                                   \\
Steps 1 2 5                & 0.492                             &  0.580        &  \textbf{0.650}    & 0.519                                                   \\
\hline
\multicolumn{5}{c}{ {\textbf{Abnormal Period: March 2017 to April 2018}}}                                                                                 \\
\textbf{Pre-processing} & \multicolumn{1}{c}{\textbf{Day}} & \multicolumn{1}{c}{\textbf{Week}} & \multicolumn{1}{c}{\textbf{Month}} & \multicolumn{1}{c}{\textbf{All periods}} \\
\hline
 Steps 1 2 3 4 5            &  0.566     & 0.476                                & 0.500                             &  0.574                           \\
 Steps 1 2 3 4              & 0.547                             & 0.444                                &  0.571     &  0.553                           \\
 Steps 1 2 3 5              & 0.531                             & 0.532                                & 0.464                             & 0.541                                                   \\
 Steps 1 2 5                &  0.584     &  0.589        & 0.429                             &  0.555    \\ \hline
\end{tabular}
\end{table}
\section{Discussion}
\label{sec:discussion}

The work carried out does not allow the exploration of all the possibilities offered by natural language processing (NLP). The approach was built iteratively, in the absence of a reference. It is based on the use of medical knowledge and  data processing.
\subsection{Strengths and Limitations}
The emergence of a problem around \Levothyrox is visible from the end of summer 2017 and this work confirms it by the only existing analysis of user comments on a single platform. Based on the statistical analysis of time series representing the frequency of words or n-grams, it turns out that many common vocabulary words of little or no relevance pollute statistical processing. It therefore becomes essential to use appropriate and codified terminologies, in order to efficiently create an extremely precise medical reference. This can help build dictionaries of common words aimed at eliminating background noise and spotting unwanted effects in a reproducible manner across all drug classes and all medical data. Classifications such as ICD-10, ATC classification should then be used. It is then a matter of bringing together under a well-designated entity all the words relating to a specific symptom, including spelling errors and the different ways of describing that symptom.

The use of basic statistical tools does not allow the early detection of abnormal events to be demonstrated. Indeed, the application of a normalization function to the daily occurrence of the Adverse Drug Reactions reported in the messages is the clear proof of the existence of an abnormal and significant event during the period when the ``scandal'' erupted in the messages. However, it does not make it possible to highlight the occurrence of abnormal events before July 2017.
It is through the use of more sophisticated computer tools that early detection of adverse events is possible. The convolutional neural network, which is a deep learning tool, effectively makes it possible to detect the first abnormalities from May 2017.
It is these tools that will improve the detection of weak signals and therefore anticipate drug problems.
The results obtained are very encouraging because there is real potential to improve them. Indeed, in this work, biases limit the ability to obtain better results and to deploy and generalize these methods. Only one drug has been studied, through messages from users of a single discussion forum. The drug in question was the subject of a case, not because of the harmful action of the active principle (by its toxic nature or its mechanism of action) but because of a change of formulation which involved communication failures. The iatrogenic effect, real for some patients but quickly resolved by dose adjustment, is not the only cause. It could be interesting and very relevant to apply this methodology on a minimum number of information sources (social networks) and specialties to compare the results obtained.
\section{Conclusion}
\label{sec:conclusion}
In this paper, we have proposed a new pharmacovigilance optimization approach by considering the use of data science and AI to collect and analyze real life data from patients. We focus on a famous drug use case: the \Levothyrox case. The new formula of \Levothyrox was marketed in France in March 2017. An increase in the frequency of Adverse Drug Reactions due to taking the drug was identified and reported in the media from July 2017. 
We develop an AI-based method for the early prediction of the undesirable effects of the drug mentioned on the endocrinology sub-forum of the \Doctissimo site. 

The first phase of the work consisted of extracting and then cleaning the data. The second phase which analyzed the best bi-grams over the period 2016 to 2020 revealed a single frequency peak between August 2017 and January 2018. The most frequent bi-gram ("Old Formula"), has a rate of appearance 160 times greater in August 2017 than in March 2017 and 6 times greater than in January 2018. “Side effects” appears 8 times in 2016, 14 times in 2019 and 183 times in 2017. The same profile is observed for other bi-grams ("New Formula", "Old Formula", "Doser Levothyrox"). The third phase of algorithmic processing was aimed at extracting and analyzing undesirable effects. In mid-September 2017, the comments showed an average frequency of occurrence of 25 per day against 6 per day between 2016 and 2020. The application of a normalization function to the curve of occurrence of adverse effects as a function of of time concluded that an unusual and significant event did indeed occur in 2017. The fourth phase made it possible, through the use of convolutional neural networks, to identify similarities between the terms used during the period of May 2017 to February 2018. Also, the algorithm confirmed that the terms cited during the period of abnormality are different from those of the period of normality. Thus, it was concluded that detection of early signals, indicators of abnormal events was possible from May 2017.

This pioneering work in the field of pharmacovigilance presents very encouraging results. It demonstrate a real capacity for innovation in the use of data science and artificial intelligence for statistical processing of real-life patient data. It makes it possible to consider, after adjusting for biases and setting up avenues for improvement, the extrapolation of the model to other data sources and other analysis scripts (with or without the use of the artificial intelligence).

 \bibliographystyle{elsarticle-num} 
 \bibliography{references}





\end{document}